\documentclass{article}
\usepackage{ijcai24}

\usepackage{times}
\usepackage{soul}
\usepackage{url}
\usepackage[hidelinks]{hyperref}
\usepackage[utf8]{inputenc}
\usepackage[small]{caption}
\usepackage{graphicx}
\usepackage{amsmath}
\usepackage{amsthm}
\usepackage{booktabs}
\usepackage{algorithm}
\usepackage{algorithmic}
\usepackage[switch]{lineno}
\usepackage{amsmath, amssymb}
\usepackage{hyperref}
\usepackage{multirow}
\usepackage{subfigure}
\usepackage{bm}

\title{Dynamic Against Dynamic: An Open-set Self-learning Framework}

\author{
}

\author{
Haifeng Yang$^{1,2}$\footnote{The first two authors contributed equally to this work}
\and
Chuanxing Geng$^{1,2,3*}$\and
Pong C. Yuen$^{3}$\And
Songcan Chen$^{1,2}$\footnote{Corresponding author}
\affiliations
$^1$Nanjing University of Aeronautics and Astronautics\\
$^2$MIIT Key Laboratory of Pattern Analysis and Machine Intelligence\\
$^3$Hong Kong Baptist University\\
\emails
\{SZ2216008, gengchuanxing, s.chen\}@nuaa.edu.cn,
pcyuen@comp.hkbu.edu.hk
}

\begin{document}

\maketitle

\begin{abstract}
    In open-set recognition, existing methods generally learn statically fixed decision boundaries using known classes to reject unknown classes. Though they have achieved promising results, such decision boundaries are evidently insufficient for universal unknown classes in dynamic and open scenarios as they can potentially appear at any position in the feature space. Moreover, these methods just simply reject unknown class samples during testing without any effective utilization for them. In fact, such samples completely can constitute the true instantiated representation of the unknown classes to further enhance the model's performance. To address these issues, this paper proposes a novel dynamic against dynamic idea, i.e., dynamic method against dynamic changing open-set world, where an open-set self-learning (OSSL) framework is correspondingly developed. OSSL starts with a good closed-set classifier trained by known classes and utilizes available test samples for model adaptation during testing, thus gaining the adaptability to changing data distributions. In particular, a novel self-matching module is designed for OSSL, which can achieve the adaptation in automatically identifying known class samples while rejecting unknown class samples which are further utilized to enhance the discriminability of the model as the instantiated representation of unknown classes. Our method establishes new performance milestones respectively in almost all standard and cross-data benchmarks.
\end{abstract}

\section{Introduction}
Traditional supervised classification has achieved great success in past decades, partly due to the closed-set assumption that the training and test samples share the same feature and label spaces \cite{fang2021learning,zhou2022open}. However, this assumption does not always hold in the real-world, since we usually know nothing about the incoming test samples, meaning they could potentially come from unknown classes unseen during training, thus drastically weakening the model performance. To meet this challenge, open-set recognition (OSR) is proposed \cite{Scheirer2013TowardOS}, which aims to construct the models that not only accurately identify known class samples but also effectively reject unknown class samples.

\begin{figure}[!t]
  \centering
  \includegraphics[width=9cm, height=4.8cm]{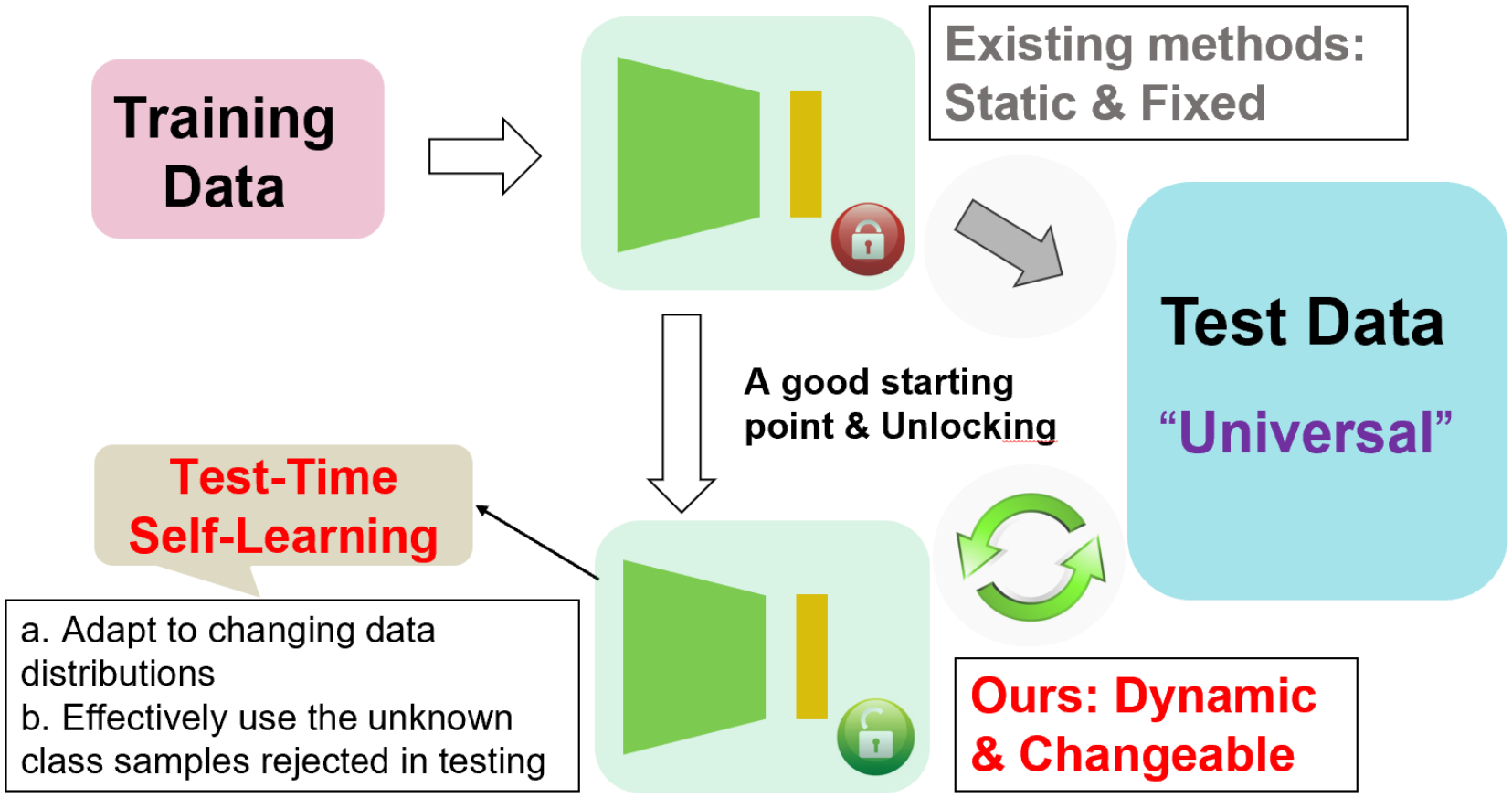}
  \caption{Differences between existing OSR methods and ours. Existing methods generally learn static and fixed classifiers using known class samples for universal unknown classes, whereas ours follows the dynamic against dynamic idea, i.e., dynamic methods against dynamic changing open-set world, where the learned classifier is dynamic and changeable.}
\end{figure}

With the unremitting efforts of the researchers, OSR has achieved significant progress \cite{Neal2018OpenSL,Oza2019C2AECC,Chen2021AdversarialRP,fang2021learning,Guo2021ConditionalVC,kong2021opengan,zhou2021learning,Xu2023ContrastiveOS}. For example, \cite{Chen2021AdversarialRP} introduced the reciprocal point technique to model the unexploited extra-class space while \cite{zhou2021learning} adopted the placeholder technique reserve space for unknown classes, thus gaining the more accurate decision boundaries. \cite{Xu2023ContrastiveOS} recently introduced the popular supervised contrastive learning technique to aim at learning more effective representations for OSR and established a new performance milestone.

Despite the encouraging advance, the ability of existing methods to deal with agnostic unknown classes actually remains limited due to the following issues:
\begin{itemize}
    \item[\checkmark] \textbf{Static and Fixed Decision Boundaries}. Existing methods generally learn statically fixed decision boundaries using known class samples (as shown in Figure 1) for the universal unknown classes in dynamic and open scenarios, which is evidently insufficient as these classes can emerge unexpectedly at any position in the feature space.
    \item[\checkmark] \textbf{Knowledge Waste of Rejected Test Samples}. Existing methods just simply reject unknown class samples without any additional operations to effectively utilize them. In fact, such samples completely can constitute the true instantiated representation of the unknown class space to further enhance the model's performance.
\end{itemize}

To address these unsolved issues in open-set recognition, this paper proposes a novel \textit{dynamic against dynamic} idea, i.e., \textit{dynamic methods against dynamic changing open-set world}. Therefore, an open-set self-learning (OSSL) framework is correspondingly developed, which starts with a good closed-set classifier trained by known classes and then self-trains with the available tested yet commonly deprecated samples for the model adaptation during testing. Specifically, OSSL first leverages the well-trained closed-set classifier to perform the identity inference of all the test samples, where it divides the entire test set into three subsets according to the logit scores of samples obtained by the closed-set classifier. Then these subsets will be submitted to a novel self-matching module which adaptively implements the classifier update. Afterwards, the updated classifier serves as the new inferrer, repeating the process until the model converges. In summary, our contributions can be highlighted as follows:
\begin{itemize}
    \item To our best knowledge, it is the first time that the dynamic against dynamic idea, i.e., dynamic methods against dynamic changing open-set world, is introduced in the modeling of the OSR problem, where an open-set self-learning framework is correspondingly developed.
    \item A novel self-matching module is design for OSSL, which can achieve the adaptation in automatically identifying known class samples while rejecting unknown samples which are further utilized to enhance the discriminability of the model as the true instantiated representation of the unknown class space.
    \item Extensive experiments verify the effectiveness of our OSSL, establishing new performance milestones respectively in almost all standard and cross-data benchmark datasets.
\end{itemize}

\section{Related Work}
\subsection{ Open Set Recognition}
The OSR problem is initially formalized in \cite{Scheirer2013TowardOS} and had achieved notable advancements in the past decade, where numerous OSR methods have been developed, which can be mainly divided into two categories: discriminative-based and generative-based methods \cite{geng2021recent}. \textbf{Discriminative-based methods} aim to accurately model known classes using various techniques in order to effectively reject unknown classes \cite{geng2020guided,Chen2020LearningOS,Chen2021AdversarialRP,zhou2021learning,huang2022class,wang2022openauc,cevikalp2023anomaly,Xu2023ContrastiveOS}. \textbf{Generative-based methods} usually employ adversarial generation techniques to generate pseudo-unknown class samples, then transforming the original $K$-class OSR problem into a $K+1$-class closed-set multi-class classification problem \cite{Neal2018OpenSL,Perera2021GeometricTN,kong2021opengan,geng2022collective}.

As mentioned previously, though these methods have shown promising results, their decision boundaries are all static and fixed, which evidently confines the further performance improvement when facing dynamic and open scenarios. This motivates us to propose the dynamic against dynamic idea, constructing dynamic changeable models to face the dynamic open-set world.

\subsection{Self-Training}
Self-training \cite{ScudderProbabilityOE,Lee2013PseudoLabelT} is a commonly used semi-supervised learning paradigm that allows the model to leverage the abundant unlabeled data to enhance its learning capabilities.  The main focus of self-training lies in that limited labeled data often make the learned model unreliable, resulting in significant noise in pseudo-labeled data, which further induces the model to update in the wrong direction. To address this issue, large numbers of methods have been proposed \cite{zou2019confidence,mukherjee2020uncertainty,xie2020self,wei2021crest,Karisani2023NeuralNA,Garg2023ComplementaryBO}. For example, \cite{zou2019confidence} proposed a confidence regularized self-training framework, treating pseudo-labels as continuous latent variables jointly optimized by alternating optimization. Recently, some researchers have attempted to extend self-training to open scenarios, i.e., there are new classes in unlabeled data that do not appear in labeled data \cite{cao2022open,Zhuang2024CredibleTF}.

Similarly, we also explore the application of self-traininig in open-set recognition. Additionally, we have relatively sufficient labeled known class samples here, which to some extent ensures the reliability of the initial classifier inference.

\subsection{Test-Time Adaptation}
Test-time adaptation (TTA) aims to improve the model performance through model adaptation with the changing test data distributions. One pipeline of this study is to jointly train a source model via both supervised and self-supervised objectives, then adapting the model through self-supervised objectives during testing \cite{sun2020test,liu2021ttt++,bartler2022mt3}. The other pipeline is the full test-time adaptation method, which adapts a model with only test data \cite{schneider2020improving,wang2020tent,zhang2022memo,iwasawa2021test,niu2022efficient}. For example, \cite{wang2020tent} proposed to adapt by test entropy minimization while \cite{zhang2022memo} adopted the prediction consistency maximization over different augmentations.

Our work falls into the latter pipeline. In particular, to our best knowledge, it is the first time to apply the TTA paradigm to the OSR scenario, further enriching the research on TTA.

\section{Methodology}
\subsection{Preliminary}
Let $\mathcal D_{tr}=\{(\textbf x_i,y_i)\}_{i=1}^{N_{tr}}$ denote the training set, where $\textbf x_i$ is the $i-$th sample and $y_i \in \mathcal C_{tr}=\{1,2,3,...,K\}$ represents the corresponding known class label. Similarly, let $\mathcal D_{te}=\{(\textbf x_j,y_j)\}_{j=1}^{N_{te}}$ be the test set, where $y_i \in \mathcal C_{te}=\{1,2,3,..,K,K+1\}$. Note that $K+1$ demonstrates a group of unknown classes, which may contain more than one class \cite{zhou2021learning}. Open-set recognition aims to learn a robust classifier that only accurately classify the known classes but also effectively reject unknown classes.

\begin{figure}
\centering
\includegraphics[width=0.49\textwidth]{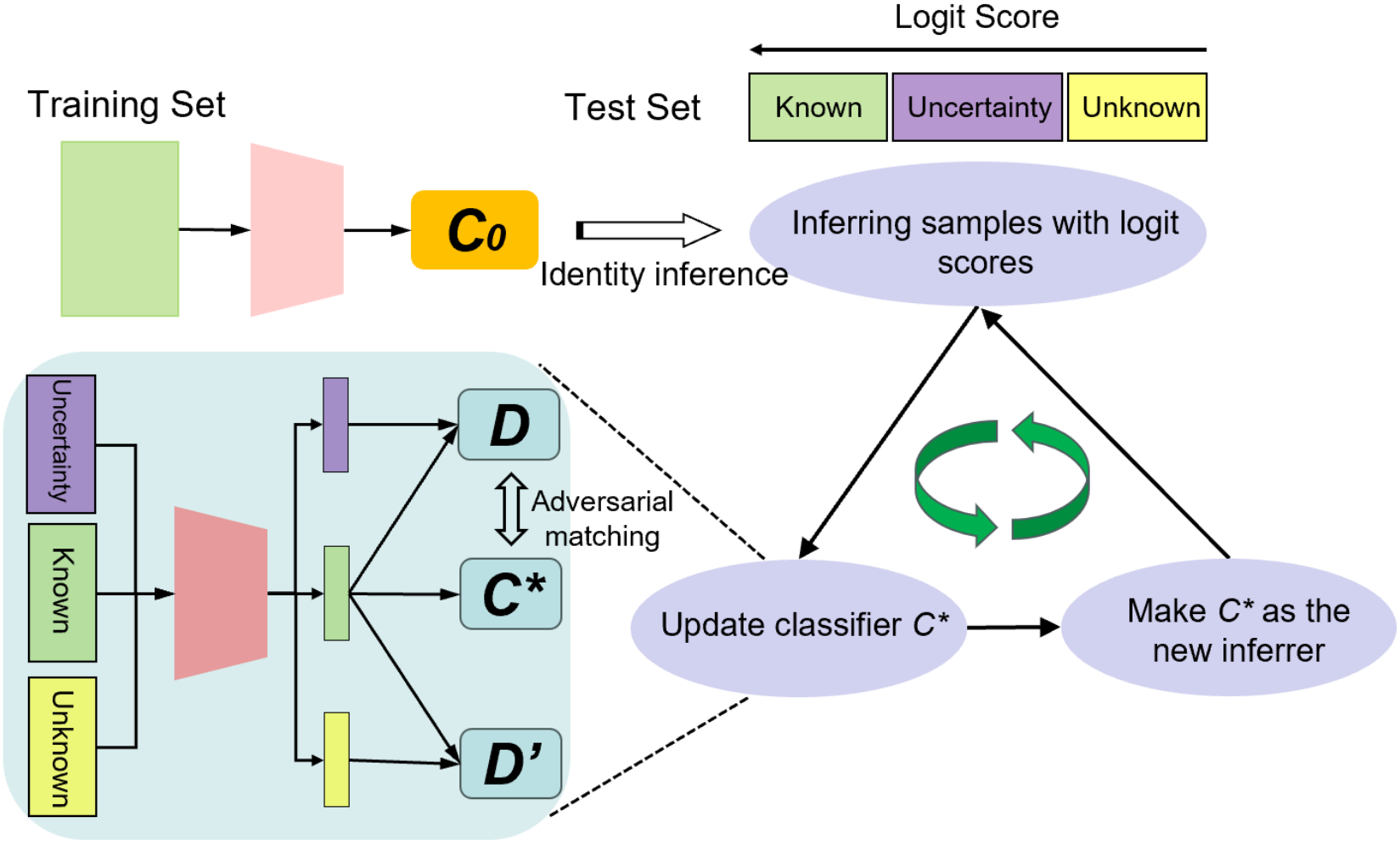}
\caption{The overview of our open-set self-learning framework.}
\label{fig2:env}
\end{figure}

\subsection{Open-set Self-learning Framework}
To face the challenge from the universal unknown classes in dynamic and open scenario, we propose the dynamic against dynamic idea and develop an open-set self-learning (OSSL) framework, which consists of two key components: one is a well-trained closed-set classifier as its starting point, and the other is a novel self-matching module. The former aims to ensure the reliability of pseudo labeling of test data to some extent, while the latter aims to achieve the model adaptation on test data. In specific, OSSL first leverages the well-trained closed-set classifier to perform the identity inference of all the test samples, where it divides the entire test set into three parts, namely known-label set, uncertainty set and unknown-label set according to the logit scores of samples obtained by the closed-set classifier. Then these data sets will be submitted to a novel self-matching module, which contains three parts: a classifier part $C^*$ applied to the known-label set, an adversarial matching part $D$ applied to the known-label and uncertainty sets, and a detection part $D'$ applied to the known-label and unknown-label sets. These parts work collaboratively to facilitate the enhancement and update of the classifier $C^*$. Afterwards, the updated classifier serves as the new inferrer, repeating the process until the model converges. Figure 2 shows the overview of our OSSL framework. Next, we will elaborate on these two key components of OSSL, i.e., the good starting point and the self-matching module.

\begin{algorithm}[tb]
    \footnotesize
    \caption{Training Procedure of OSSL Framework}
    \label{alg:algorithm}
    \textbf{Input}:
    Data from training set $\mathcal D_{tr}$;
    Data from testing set $\mathcal D_{te}$; \\
    \textbf{Parameter}:
    Starting point classifier $C_0(\cdot)$;
    Adversarial matching part $D(\cdot)$;
    Detection part $D'(\cdot)$;
    \begin{algorithmic}[1] 
        \STATE Initialize model parameters $D(\cdot)$, $D'(\cdot)$ .
        \FOR{$epoch$ =1 in $epoch_{max}$}
        \STATE Divide  $\mathcal D_{te}$ based on probability scores from classifier $C^*$ (its initial one is $C_0$) into three parts:
        $T_1$ , $T_2$ and $T_3$.
        \STATE Sample data from training set to get  $Tr_s \subset \mathcal D_{tr}$.
        \STATE Based on Eq.(6) , using data from $Tr_s$ and known-label set $T_1$ to calculate $\mathcal{L}_{C^*}'$ .
        \STATE Obtain data from known-label set $T_1$, unknown-label set $T_3$ and $Tr_s$ to compute $\mathcal{L}_{D'}'$ based on Eq.(7) .
        \STATE Calculate $\omega^s$ and $\omega^t$ in the method of Eq.(3) .
        \STATE Based on Eq.(8) , utilize calculated $\omega^s$ and $\omega^t$ to compute $\mathcal{L'}_D$ for our adversarial matching part.
        \STATE Based Eq.(9) , use data from unknown-label set $T_3$ to calculate $\mathcal{L}_{Mar}$.
        \STATE Calculate the total loss of our OSSL framework based on Eq.(10). \STATE Based on the loss function, update parameters of our framework.
        \ENDFOR
    \end{algorithmic}
\end{algorithm}

\subsubsection{The Good Starting Point of OSSL}
The core of this component is that the inference of initial classifier should be reliable to some extent, especially for unknown classes. Fortunately, the recent study  \cite{Vaze2022OpenSetRA} has shown that a well-trained closed-set classifier can achieve this, where they just train a closed-set classifier using various tricks such as data augmentation, label smoothing, etc, achieving at least comparable performance to SOTA baselines \cite{Chen2021AdversarialRP,Xu2023ContrastiveOS}. Therefore, we here employ the network architecture in \cite{Vaze2022OpenSetRA} as our model's backbone, the trained closed-set classifier as our initial classifier $C_0$ whose outputs are logit vectors.

The role of the initial classifier $C_0$ is to provide the first identity inference for the test data, where it divides the test set into three parts according the logit scores of samples obtained by $C_0$, namely, known-label set $T_1 = \{\bm{x}_j|\max(C_0(\bm{x}_j))>
\mu\}$, unknown-label set $T_3 = \{\bm{x}_j|\max(C_0(\bm{x}_j))-\min(C_0(\bm{x}_j))<\gamma\}$, and the remaining data set, i.e., uncertainty set $T_2$.  Note that considering the characteristic of the unknown class logit vectors, which tends to be uniformly distributed, we here adopt the relative differences of logit values rather than their magnitudes. Then these preliminarily inferred samples will be submitted to the self-matching module of OSSL for further processing.

\textbf{Remark}. In fact, all existing OSR approaches can serve as our initial classifiers. This implies that our OSSL is orthogonal to existing methods and can leverage them as starting points to achieve even better performance.

\subsubsection{The Self-matching Module of OSSL}
The self-matching module aims to efficiently update the classifier $C^*$ to adapt to the changing test distribution. In this module, the known-label set $T_1$ is directly used to update $C^*$, however, the number of its samples may not be sufficient, just about 28\% of the batch-size (256) in our framework. Thus, the adversarial matching part $D$ is proposed to fully utilize the uncertainty set $T_2$ to update $C^*$, whose core is a sample-level weighted mechanism, mainly drawing on the idea of distribution alignment in \cite{You2019UniversalDA}. $D$ aims to maximize the alignment/matching of known-class data distribution in $T_1$ and $T_2$ while automatically detecting unknown class samples in $T_2$. Further, the key to the sample-level weighted mechanism here is to quantify the similarity between each sample and the known class distribution, which can be achieved by the detection part $D'$ as its objective is to predict samples from known class distribution as 1 and samples from unknown class distribution as 0. In summary, these three parts work collaboratively to facilitate the enhancement and update of $C^*$. Their specific details will be elaborated in the followings.

For the classifier part, it mainly applies the cross-entropy loss to known-label set to gain the classifier update,
\begin{equation}
    \mathcal{L}_{C^*}=
    \mathbb{E}_{\bm{x}\sim T_1}L(\widehat{\bm{y}},\text{softmax}(C^*(F(\bm{x}))),
\end{equation}
where $F(\cdot)$ denotes the feature extractor network, $C^*$ is a linear fully connected layer classifier with only one layer, whose initialization parameters come from $C_0$ and outputs are logit vectors. $\widehat{\bm{y}}$ denotes the pseudo-known label obtained by the previous classifier (the initial one is $C_0$).

For the adversarial matching part, it operates on known-label and uncertainty sets, defined as follows:
\begin{equation}
    \mathcal{L}_{D}= -\mathbb{E}_{\bm{x}\sim T_1} \omega^{s}\log(D(F(\bm{x}))) - \mathbb{E}_{\bm{x}\sim T_2}
    \omega^{t}\log(1-D(F(\bm{x}))).
\end{equation}
Its core is the sample-level weighted mechanism, i.e., $\omega^{s}$ and $\omega^{t}$, which represent the probabilities of corresponding samples belonging to known classes, and their values can be obtained by the following equations,
\begin{equation}
    \omega^{s}(\bm{x})=\frac{H(\widehat{\bm{y}})}{\log|\mathcal{C}_{tr}|}-d', \ \ \omega^{t}(\bm{x})=d'-\frac{H(\widehat{\bm{y}})}{\log|\mathcal{C}_{tr}|},
\end{equation}
where $H$ denotes the entropy of a vector, which is normalized by its maximum value
(${log|\mathcal{C}_{tr}|}$) so that $\omega^{s}$ and $\omega^{t}$ are restricted into [0, 1].  $|\mathcal{C}_{tr}|$ represents the number of known classes. $d'$ quantifies the similarity of $\bm{x}$ to known classes and can be achieved by the detection part $D'$ ($d'=D'(F((\bm{x})))$), defined as follows:
\begin{equation}
     \mathcal{L}_{D'}= - \mathbb{E}_{\bm{x}\sim T_1}\log(D^{'}(F(\bm{x}))) - \mathbb{E}_{\bm{x}\sim T_3}\log(1-D^{'}(F(\bm{x}))) .
\end{equation}

In summary, the total loss function of the self-matching module can be written as:
\begin{align*}
    \max \limits_D \min\limits_{F, C^*} &\ \mathcal{L}_{C^*}-\mathcal{L}_{D}
\end{align*}
\begin{align}
    \min\limits_{D^{'}}& \ \mathcal{L}_{D'}
\end{align}

\subsubsection{Some Enhancement Strategies}
To further improve the reliability of model inference, we enhance the model from the following two strategies.

\textbf{Injecting a small amount of ground-truth data}. We sample a small amount of labeled data of known classes from the training set, i.e. $Tr_s \subset \mathcal D_{tr}$, and add it to the self-learning process
of the model to enhance the reliability of model inference. As a result, the loss functions of the self-matching module respectively become:

\begin{equation}
\begin{aligned}
\mathcal{L}_{C^*}'
&=\mathbb{E}_{\bm{x}\sim T_1} L(\widehat{\bm{y}},\text{softmax}(C^*(F(\bm{x})))  \\
& +\mathbb{E}_{\bm{x}\sim Tr_s}L(\bm{y},\text{softmax}(C^*(F(\bm{x}))),
\end{aligned}
\end{equation}

\vspace{0.3cm}

\begin{equation}
\begin{aligned}
  \mathcal{L}_{D'}' =
  &- \mathbb{E}_{\bm{x}\sim T_1 \cup Tr_s}\log(D^{'}(F(\bm{x})))  \\
  &-\mathbb{E}_{\bm{x}\sim T_3}\log(1-D^{'}(F(\bm{x}))),\\
\end{aligned}
\end{equation}

\vspace{0.3cm}

\begin{equation}
\begin{aligned}
  \mathcal{L}_{D}' =
  &-\mathbb{E}_{\bm{x}\sim T_1 \cup Tr_s} \omega^{s}\log(D(F(\bm{x}))) \\
  &- \mathbb{E}_{\bm{x}\sim T_2}
    \omega^{t}\log(1-D(F(\bm{x}))).\\
\end{aligned}
\end{equation}

\textbf{Marginal logit loss for unknown classes}. In fact, we usually expect the logits of unknown class samples to tend towards the uniform distribution. However, such uniformly-distributed logits do not necessarily lead to sufficiently small values of them. To address this issue, we here introduce a marginal logit loss for the unknown-label set $T_3$ inspired by \cite{deng2022learning}, defined as follows,
\begin{align}
    \mathcal{L}_{Mar}=\frac{1}{N_{T_3}}\sum_{i=1}^{N_{T_3}} \sum_{k=1}^{K}max(0,\lambda + v_{i,k}),
\end{align}
where $v_{i,k}$ denotes the element value to the corresponding position of the logit vector. Therefore, the ultimately total loss of our OSSL can be summarized as follows:

\begin{align*}
    \max \limits_D \min\limits_{F, C^*} &\ \mathcal{L'}_{C^*}-\mathcal{L'}_{D}
\end{align*}
\begin{align}
    \min\limits_{D^{'}, F}& \ \mathcal{L'}_{D'} + \mathcal{L}_{Mar}.
\end{align}
The overall training procedure of OSSL is detailed in Algorithm 1.


\begin{table*}
\centering
\footnotesize
\tabcolsep 1.8mm
\renewcommand\arraystretch{1.2}
\begin{tabular}{lcccccc}
\hline
\textbf{Method} & \textbf{MNIST} & \textbf{SVHN}& \textbf{Cifar10}& \textbf{Cifar+10}& \textbf{Cifar+50}&\textbf{TinyImageNet}\\
\hline
Softmax & 0.978  &0.886 &0.677& 0.816&0.805&  0.577 \\
OpenMax (CVPR'2016) & 0.981  &0.894 &0.654& 0.817&0.795&  0.576 \\
OSRCI (ECCV'2018) & 0.988  &0.910 &0.699& 0.838&0.827&  0.586 \\
CROSR (CVPR'2019)& 0.991  &0.899 &0.883& 0.912&0.905&  0.589 \\
C2AE (CVPR'2019)& 0.989  &0.922 &0.895& 0.955&0.937& 0.748 \\

OpenHybrid (ECCV'2020) & 0.995  &0.947 &0.950& 0.962&0.955&  0.793 \\
PROSER (CVPR'2021) & -  &0.943 &0.891& 0.960&0.953&  0.693 \\
EGT (ICME'2021) & -  &0.958 &0.821& 0.937&0.930& 0.709 \\
ARPL (TPAMI'2021) & 0.996  &0.963 &0.901& 0.965&0.943&  0.762 \\
ARPL+CS (TPAMI'2021) & 0.997  &0.967 &0.910& 0.971&0.951& 0.782 \\
DIAS (ECCV'2022) &0.992 &0.943&0.850&0.920&0.916&0.731\\
PMAL (AAAI'2022) &0.997 &0.970&0.951&0.978&0.969&0.831\\
ALL\_U\_NEED (ICLR'2022) & 0.993  &0.971 &0.936& 0.979&0.965&  0.830 \\
ConOSR (AAAI'2023)&0.997&\textbf{0.991}&0.942&0.981&0.973&0.809\\
\hline
ARPL+Ours & 0.997 (\textbf{+0.1\%})  &0.969 (\textbf{+0.6\%}) &0.919 (\textbf{+1.8\%})& 0.972 (\textbf{+0.7\%})&0.951 (\textbf{+0.8\%})&  0.772 (\textbf{+1.0\%})\\
OSSL (Ours) & \textbf{0.998}  &0.976  &\textbf{0.952} & \textbf{0.988}  &\textbf{0.980}  &  \textbf{0.842}\\
\hline
\end{tabular}
\caption{\label{citation-guide}
Evaluation on open-set detection (AUROC) under the standard-dataset setting. APRL+Ours means that the classifier in ARPL is chosen as the starting point classifier.  All of the reported results displayed are averaged across the same five distinct splits.}
\end{table*}

\section{EXPERIMENTS}
\subsection{Implementation Details}
We employ the network architecture in \cite{Vaze2022OpenSetRA} as the backbone of our learning framework\footnote{\url{https://github.com/ChuanxingGeng/OSSL}}, and choose the Stochastic Gradient Descent (SGD) technique as the optimizer. The same data augmentations in \cite{Vaze2022OpenSetRA}, like random horizontal flip, random cropping, etc., are also adopted here. Considering that the feature extractor $F(\cdot)$ is a already well-trained network, we here set its learning rate to $10^{-4}$ for TinyImageNet while $10^{-5}$ for other datasets. Furthermore, the learning rates of other parts except for $F(\cdot)$ are uniformly set to 0.01. For the threshold parameters used in the partition of test set, we set $\mu = 0.3,\ \gamma = 0.03$ for TinyImageNet, $\mu = 0.5,\ \gamma = 0.03$ for Cifar10, Cifar+10, Cifar+50, while $\mu = 0.8,\ \gamma = 0.02$ for MNIST and SVHN. In addition, the number of samples from $Tr_s$ in each batch is set to 16 (the batch-size is 256), while the hyper-parameter $\lambda$ in $\mathcal{L}_{Mar}$ is set to 2 for all benchmark datasets.

\subsection{Experiments in Standard-Dataset Benchmarks}
\subsubsection{Datastes} We here follow the protocol defined in \cite{Neal2018OpenSL}, and provide six standard OSR benchmarks:
\begin{itemize}
    \item \textbf{MNIST, SVHN, Cifar10}. For MNIST \cite{Lake2015}, SVHN \cite{Netzer2011}, and CIFAR10 \cite{Krizhevsky2009}, each dataset comprises 10 distinct classes, in which 6 classes are randomly selected as known classes and the other 4 classes as unknown.
    \item \textbf{Cifar+10, Cifar+50}. In this series of experiments, we select 4 classes from the CIFAR10 dataset to serve as known classes for training and 10 or 50 classes from the CIFAR100 dataset  as unknown classes in specific experiment.
    \item \textbf{TinyImageNet}. TinyImageNet, a derived subset of the larger ImageNet \cite{Russakovsky2014ImageNetLS} dataset, encompasses 200 classes, where 20 classes are selected as known classes, while the remaining 180 classes as unknown classes.
\end{itemize}

\subsubsection{Evaluation Metrics}
Following the mainstream works \cite{geng2021recent}, the area under ROC curve (AUROC) is used to evaluate the model's ability to detect unknown classes shown in Table 1, while the accuracy (ACC) is used to evaluate the performance of closed-set classification detailed in the supplementary materials.



\subsubsection{Results Comparison}
We compare OSSL with 14 classical and leading OSR methods including Softmax, OpenMax \cite{Bendale2015TowardsOS}, OSRCI \cite{Neal2018OpenSL}, CROSR \cite{Yoshihashi2018ClassificationReconstructionLF}, C2AE \cite{Oza2019C2AECC},  OpenHybrid \cite{Zhang2020HybridMF}, PROSER \cite{zhou2021learning}, EGT \cite{Perera2021GeometricTN}, ARPL \cite{Chen2021AdversarialRP}, DIAS \cite{Moon2022DifficultyAwareSF},
PMAL \cite{Lu2022PMALOS} , All$\_$U$\_$NEED \cite{Vaze2022OpenSetRA} and
\cite{Xu2023ContrastiveOS} ConOSR.


Table 1 reports the AUROC results. Thanks to the self-matching module in testing endowing the model with the adaptability to the changing data distributions, our OSSL establishes new performance milestone on almost all benchmark datasets. In particular, in some relatively difficult classification tasks, like TinyImageNet, it significantly outperforms the current two SOTA baselines, i.e., ALL\_U\_NEED \cite{Vaze2022OpenSetRA} and ConOSR \cite{Xu2023ContrastiveOS}, respectively achieving performance improvements of 1.2 and 3.3 percentage points.

To further demonstrate the effectiveness of our learning framework, we also employ a popular OSR method, i.e., ARPL \cite{Chen2021AdversarialRP} as the starting point of our OSSL (APRL+Ours). As shown in Table 1, APRL+Ours achieve performance gains across all benchmark datasets. For instance, it achieves an average improvement of $0.35\%$ on MNIST and SVHN whose performance is almost saturated for APRL, while an average improvement of about $1.1\%$ on the remaining benchmark datasets. This also implies that our OSSL is orthogonal to existing methods and can assist them in breaking through existing limitations, achieving further performance improvement.

\subsection{Experiments in Cross-Dataset Benchmarks}
In this series of experiments, we adhere to the established protocol in \cite{Yoshihashi2018ClassificationReconstructionLF} to execute our experiments. Within this protocol, all the classes designated for training in the original dataset are employed as in-distribution (ID) data. Meanwhile, samples from an auxiliary dataset are incorporated into the test set, serving as out-of-distribution (OOD) data.
\subsubsection{Datasets}
\begin{itemize}
    \item \textbf{ID: MNIST; OOD: Omniglot, MNIST-Noise, Noise}. MNIST is selected as ID data, while samples from Omniglot, MNIST-Noise, Noise are selected as OOD data. In specific, the Omniglot dataset consists of a diverse collection of hand-written characters from various alphabets. The Noise dataset is synthesized with each image generated by sampling pixel values from a uniform distribution in the range [0, 1]. Additionally, the MNIST-Noise dataset is created by superimposing the test images from the MNIST dataset onto the Noise images. The OOD data is quantified at 10,000 instances, which equals to the number of test samples in MNIST.
    \item \textbf{ID: Cifar10; OOD: ImageNet, LSUN}. Similarly, Cifar10 here is selected as ID data, while samples from ImageNet and LSUN are selected as OOD data. Since the image size of ImageNet and LSUN \cite{Yu2015LSUNCO} is different from Cifar10, two different ways, i.e., crop and resize operations are used to adjust their image sizes to match those in Cifar10. In addition, the numbers of test samples for Cifar10, ImageNet and LSUN are all $10,000$.
\end{itemize}

\subsubsection{Evaluation Metrics}
The performance evaluation is conducted using the macro-averaged F1-score (\%), which provides a comprehensive measure of the model's effectiveness across the diverse class types.

\begin{table}
\centering
\footnotesize
\tabcolsep 2.4mm
\renewcommand\arraystretch{1.2}
\begin{tabular}{lccc}
\hline
\textbf{Method} & \textbf{Omniglot} & \textbf{MNIST-Noise}& \textbf{Noise}\\ 
\hline
Softmax & 59.5  &64.1 &82.9\\
OpenMax  & 68.0  &72.0 &82.6 \\
CROSR  & 79.3  &82.7 &82.6 \\
PROSER  & 86.2  &87.4 &88.2 \\
ConOSR  & 95.4  &98.7 &98.8 \\
\hline
OSSL(Ours) & \textbf{97.6} (+\textbf{2.2}) &\textbf{99.7} (+\textbf{1.0})&\textbf{99.2} (+\textbf{0.2}) \\
\hline
\end{tabular}
\caption{\label{citation-guide}
Macro-F1 score (\%) of different methods under the cross-dataset setting (MNIST as the ID data). }
\end{table}

\begin{table}
\centering
\footnotesize
\small
\tabcolsep 1.0mm
\renewcommand\arraystretch{1.2}
\begin{tabular}{lcccc}
\hline
  \multirow{2}{*}{\textbf{Method}} & \multicolumn{2}{c}{\textbf{ImageNet}} & \multicolumn{2}{c}{\textbf{LSUN}} \\
                  & \textbf{Crop} & \textbf{Resize} & \textbf{Crop} & \textbf{Resize} \\
\hline
Softmax& 63.9  &65.3 &64.2& 64.7 \\
OpenMax & 66.0  &68.4 &65.7& 66.8 \\
OSRCI & 63.6  &63.5 &65.0& 64.8 \\
CROSR & 72.1  &73.5 &72.0& 74.9\\
GFROSR  & 75.7  &79.2 &75.1& 80.5 \\
PROSER  & 84.9  &82.4 &86.7& 85.6 \\
ConOSR  & 89.1  &84.3 &91.2& 88.1 \\
\hline
OSSL(Ours) & \textbf{96.5} (\textbf{+7.4}) &\textbf{90.8} (\textbf{+6.5}) &\textbf{96.4} (\textbf{+5.2})& \textbf{95.5} (\textbf{+7.4})\\
\hline
\end{tabular}
\caption{\label{citation-guide}
Macro-F1 score (\%) of different methods under the cross-dataset setting (Cifar10 as the ID data).}
\end{table}

\subsubsection{Results Comparison}
We compare our OSSL with 7 classical and leading methods. Table 2 and Table 3 respectively report the results on the MNIST and Cifar10 tasks. Compared with the SOTA baseline, i.e., ConOSR \cite{Xu2023ContrastiveOS}, our OSSL creates new performance milestones with significant advantages on all cross-dataset benchmarks. In particular, for the relative challenging Cifar10 task, it beats ConOSR with an average performance gain of about 6.6\%, while an average performance gain of about 1.2\% for the MNIST task. This once again proves the effectiveness of such an dynamic changeable learning framework for changing data distributions.


\begin{figure}[]
  \centering
  \subfigure[]{\includegraphics[width=0.48\linewidth,height=3cm]{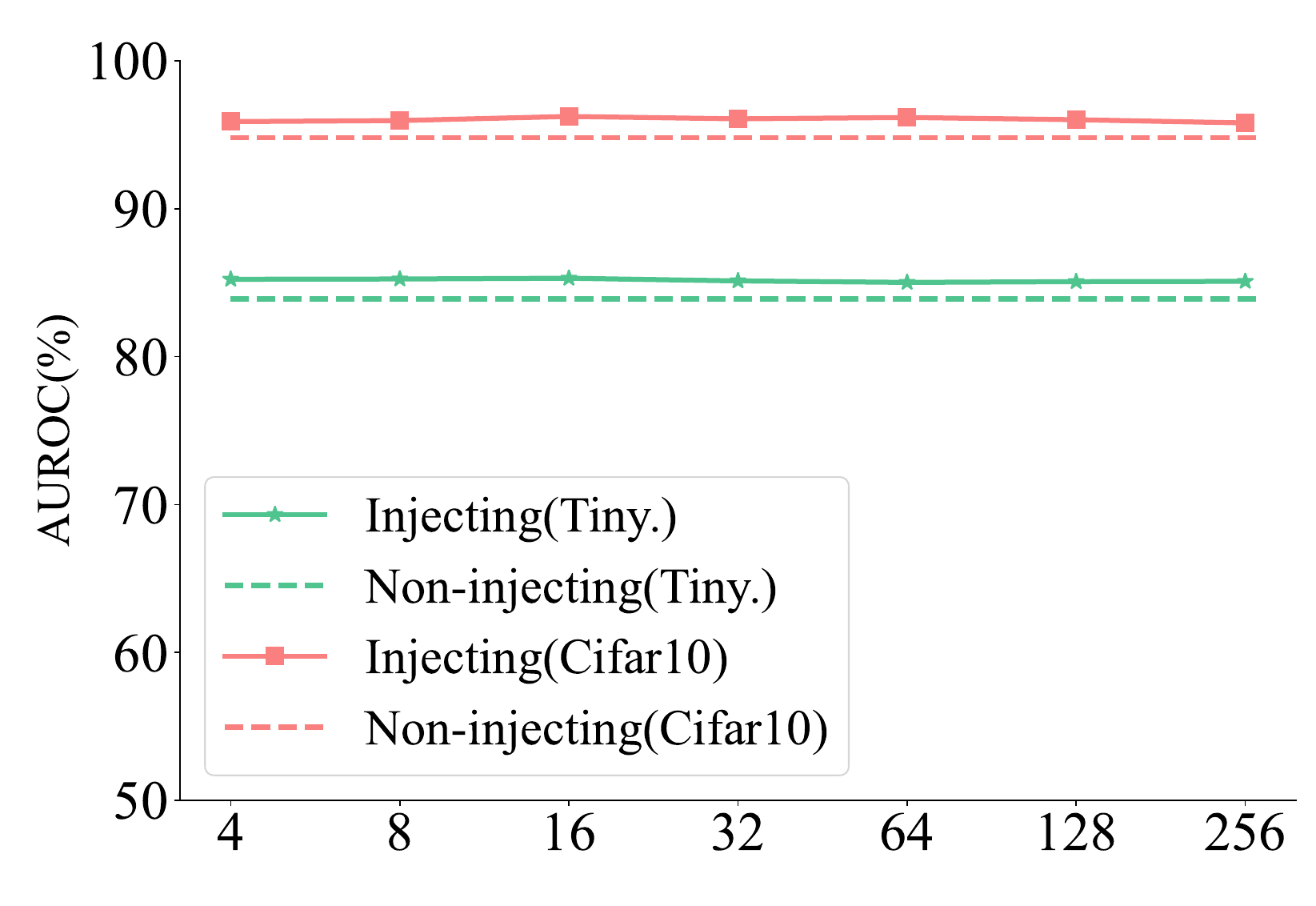}}
  \hfill
  \subfigure[]{\includegraphics[width=0.45\linewidth,height=3cm]{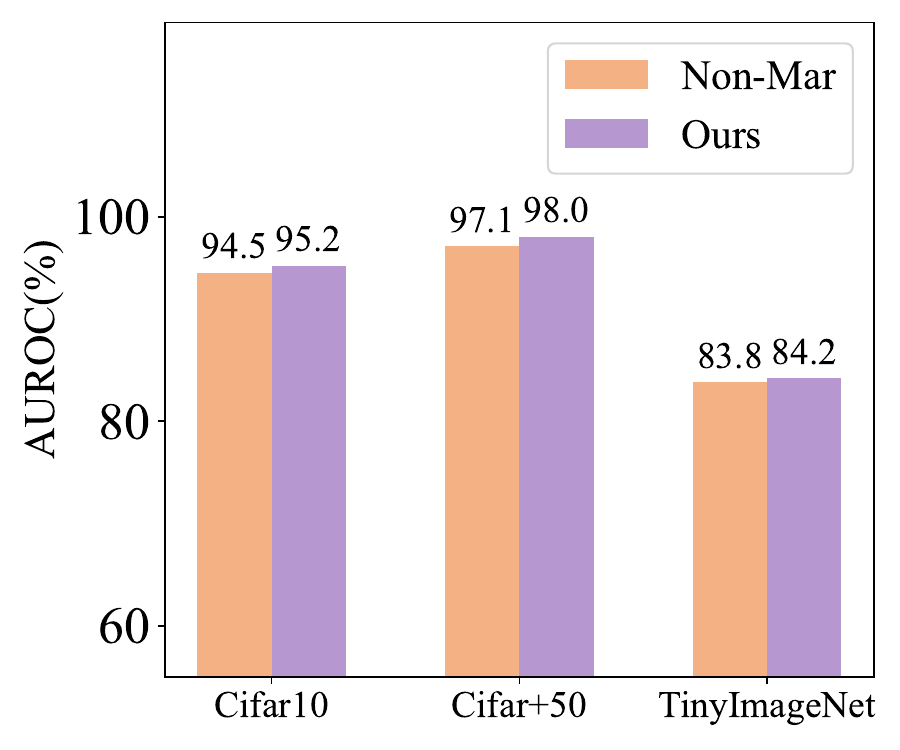}}
  \label{fig:sensitivity}
  \caption{Experiments on the effectiveness of different enhancement strategies.
  }
\end{figure}

\begin{figure}[]
  \centering
  \subfigure[]{\includegraphics[width=0.48\linewidth,height=4cm]{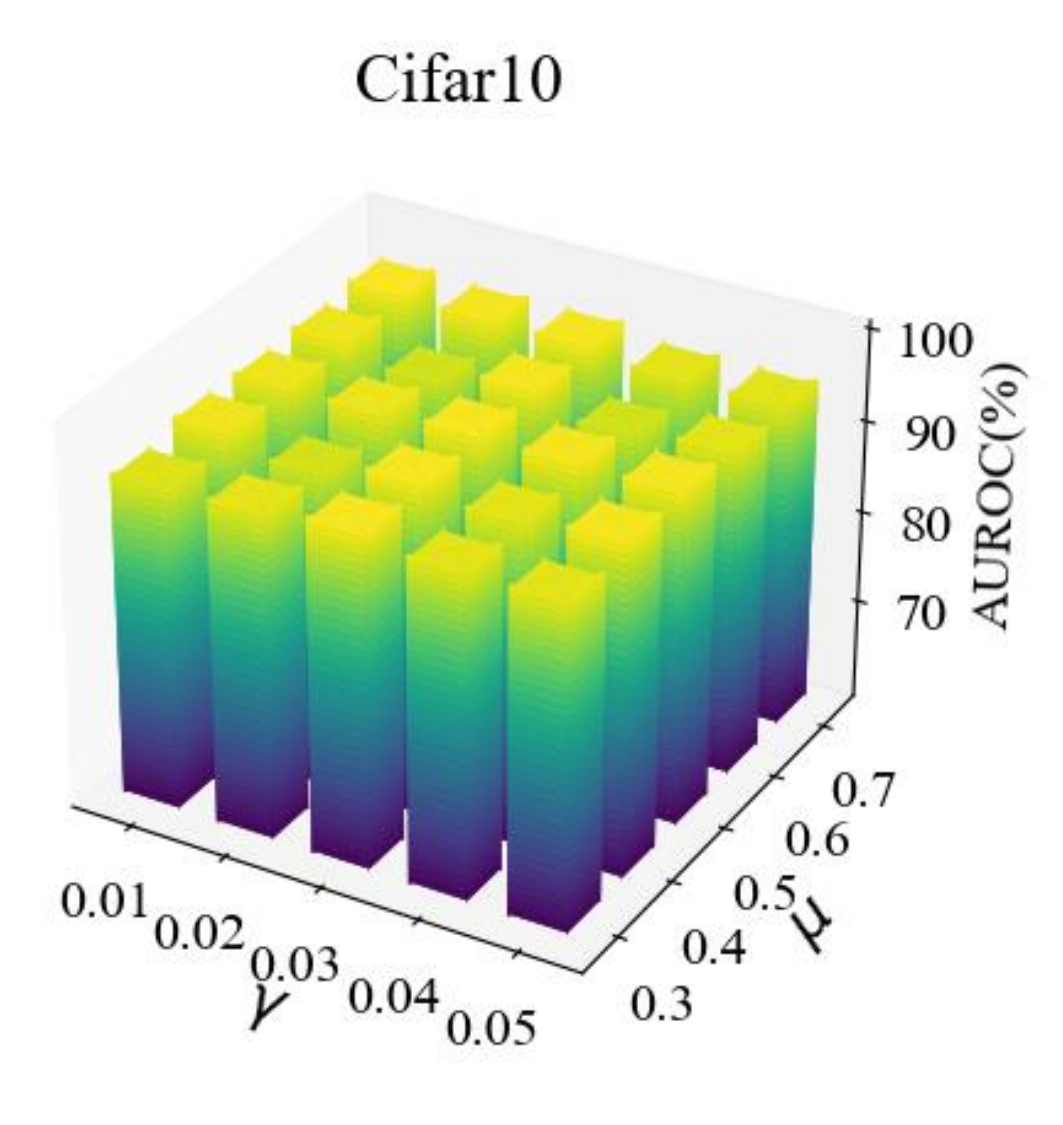}}
  \hfill
  \subfigure[]{\includegraphics[width=0.48\linewidth,height=4cm]{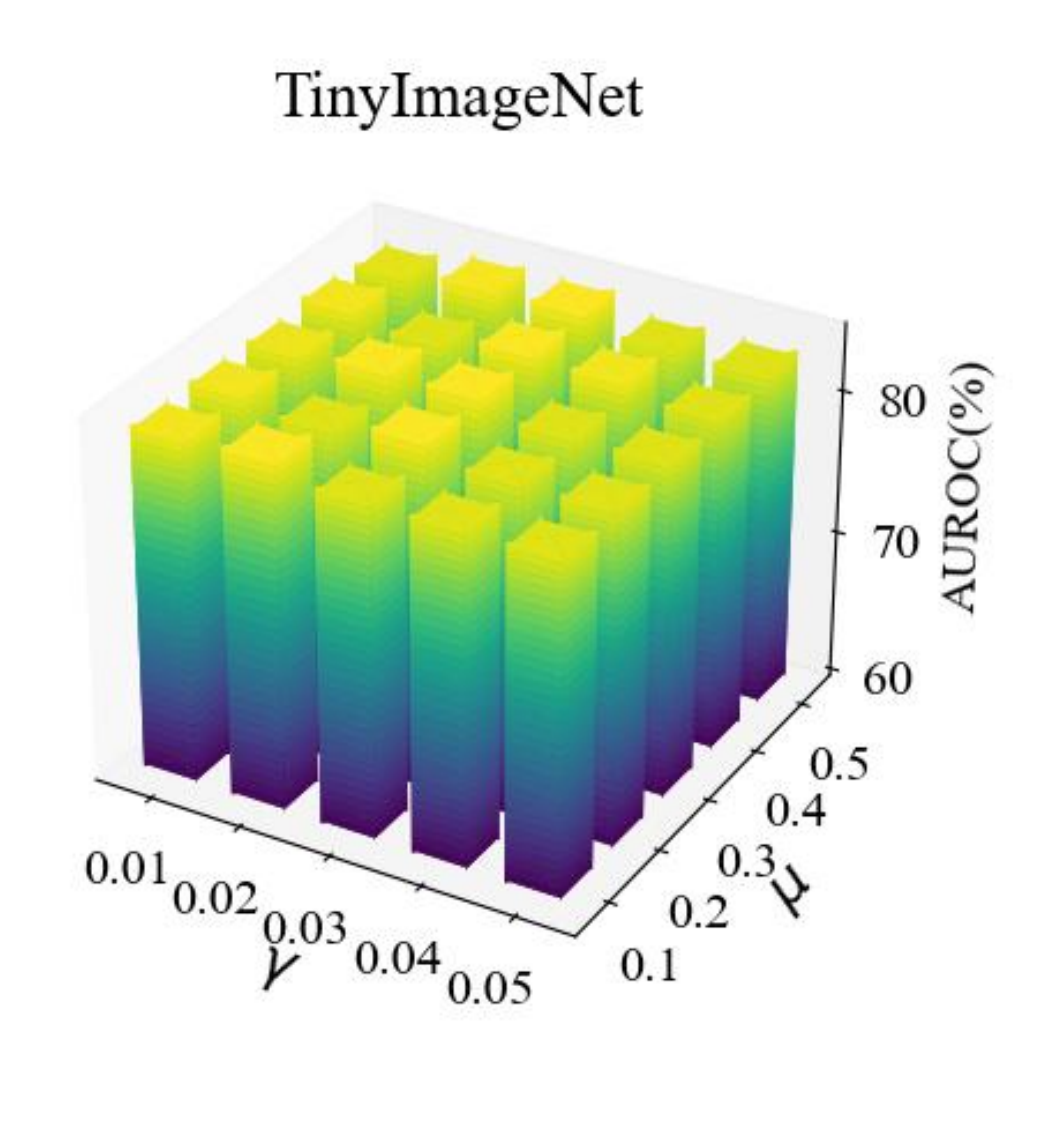}}
  \label{fig:sensitivity}
  \caption{Threshold parameter sensitivity analysis on Cifar10 and TinyImageNet, where (a) reports the results for Cifar10 while (b) for TinyImageNet.
  }
\end{figure}

\subsection{Effectiveness of Different Enhancements}
\subsubsection{Injecting a small amount of ground-truth data}
To verify the effectiveness of injecting a small amount of labeled samples, we here conduct the experiments on Cifar10 and TinyImageNet via changing the number of labeled samples from the training set in each batch (the batch size is 256 in all experiments here) during test-time training. Figure 3(a) shows the results. \textbf{Our first observation} is that whether injecting a small number of labeled samples to the Cifar10 task or the TinyImageNet task, the corresponding performances are significantly improved with about 0.4\% gains for Cifar10 while about 0.35\% gains for TinyImageNet. This verifies that adding a small number of labeled samples can improve the inference performance of the model. \textbf{Our second observation} is that as the number of labeled samples increases, there is no obvious trend of performance growth, which can be mainly attributed to the fact that the initial classifier has been well-trained on these samples. Please note that this may not necessarily be a bad thing. In fact, it provides evidence (at least to some extent) in such a scenario where we only obtain a well-trained black-box model and a small amount of authorized training samples, we can still train a model that performs comparably to models trained with large numbers of authorized training samples.

\subsubsection{Marginal logit loss for unknown classes}
To verify the effectiveness of the introduced marginal logit loss for unknown classes, we also conduct the relevant experiments. Due to the limited space, we here only present the results on Cifar10, Cifar+50, and TinyImageNet (Figure 3(b)), while the results on other datasets can be found in the supplementary materials. As shown in Figure 3(b), the introduce of $\mathcal{L}_{Mar}$ enables the model to achieve consistent performance improvement on all benchmark datasets.

\begin{figure}
\centering
\includegraphics[width=0.42\textwidth]{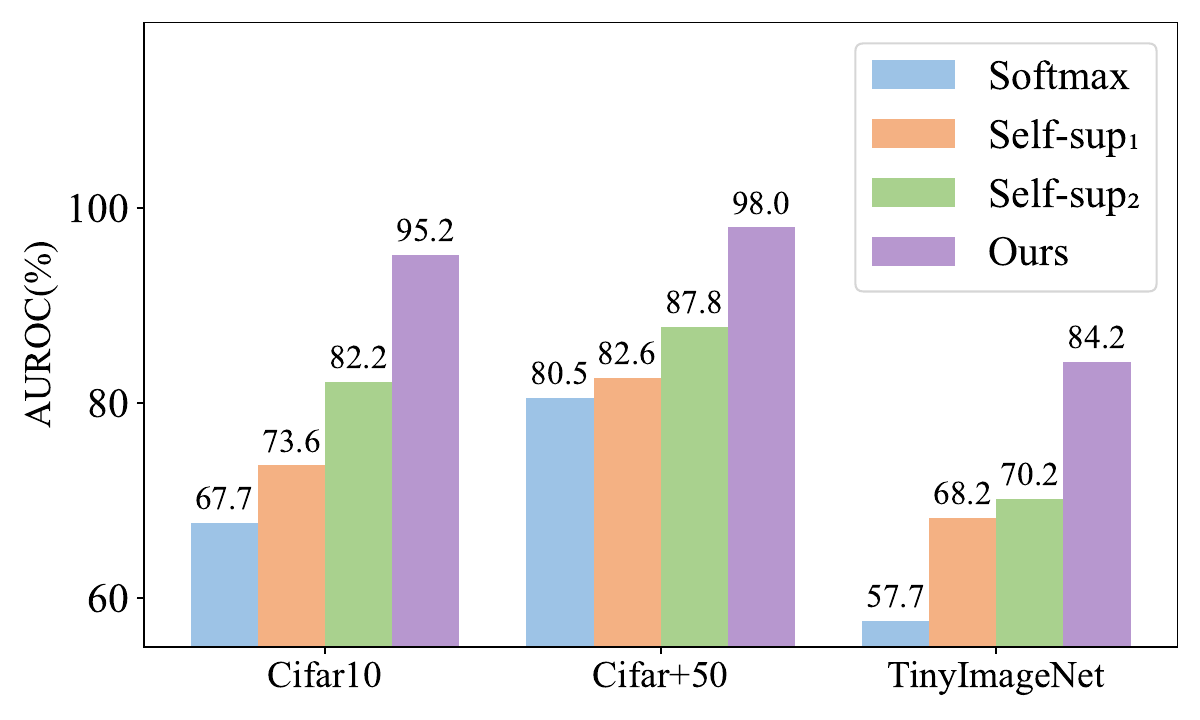}
\caption{Performance comparisons between our OSSL and two self-supervised learning methods.}
\label{fig1:env}
\end{figure}

\subsection{Threshold Parameter Sensitivity Analysis}
In this section, we undertake a specific sensitivity analysis focusing on the threshold parameter associated with the partition of the test set. We have two threshold parameters, i.e., $\gamma$ and $\mu$. Limited by the space, we here take the experiments on Cifar10 and TinyImageNet as the examples. In specific, For the parameter $\gamma$, we change its value in the range of \{0.01, 0.02, 0.03, 0.04, 0.05\} both for Cifar10 and TinyImageNet. For the parameter $\mu$, we change its value in the range of \{0.3, 0.4, 0.5, 0.6, 0.7\} for Cifar10 while  \{0.1, 0.2, 0.3, 0.4, 0.5\} for TinyImageNet. Figure 4 shows the results.

As shown in Figure 4, our model is relatively insensitive to the choice of $\gamma$ and $\mu$ whether on Cifar10 or TinyImageNet. On one hand, it indicates that the selected closed-set classifier in our OSSL has already possessed a significant level of inference capability to some extent. On the other hand, it also provides evidence that we can select an effective threshold without the need for excessive parameter searching. Still, careful selection of these parameters would result in better performance.

\subsection{Comparison with Self-Supervised Learning}
To further demonstrate the advantages of our learning framework, we here also compare our OSSL with the self-supervised learning method which can use the unlabeled test data as well. In specific, two self-supervised learning methods for utilizing unlabeled test data are provided. One adopts a joint training way ($\text{Self-sup}_1$), where the cross-entropy (CE) loss is applied to the labeled training set, and the labeled training set is combined with the unlabeled test set to form a unified unlabeled dataset that is fed into the self-supervised loss (the contrastive loss here). Then these two losses work together to guide the learning of the network. The other one employs a two-stage training way ($\text{Self-sup}_2$). In the first stage, the unlabeled dataset composed of the training and test sets is fed into the self-supervised loss to train the network. In the second stage, building upon the model trained in the first stage, the entire network is fine-tuned using the labeled training set with the cross-entropy loss. For the sake of fairness, all these methods, like ours, employ the network architecture in \cite{Vaze2022OpenSetRA}. Limited by the space, we here just provide the results in Cifar10, Cifar+50, TinyImageNet, while for the results on the remaining datasets, please refer to the supplementary materials.

As shown in Figure 5, though the performances of these two self-supervised methods have significantly improved compared to the baseline method, i.e., Softmax, on all benchmark datasets, there is still a significant gap between their performance and our OSSL. We believe that this is mainly attributed to the following reasons: i) There may be some conflicts between CE loss and self-supervised loss during training, thus interfering the learning of the network; ii) The self-supervised proxy tasks may mismatch with the downstream classification task to some extent; iii) The learned features may be too generalized, which is not conducive to the detection of unknown classes.

\begin{figure}
\centering
\includegraphics[width=0.4\textwidth]{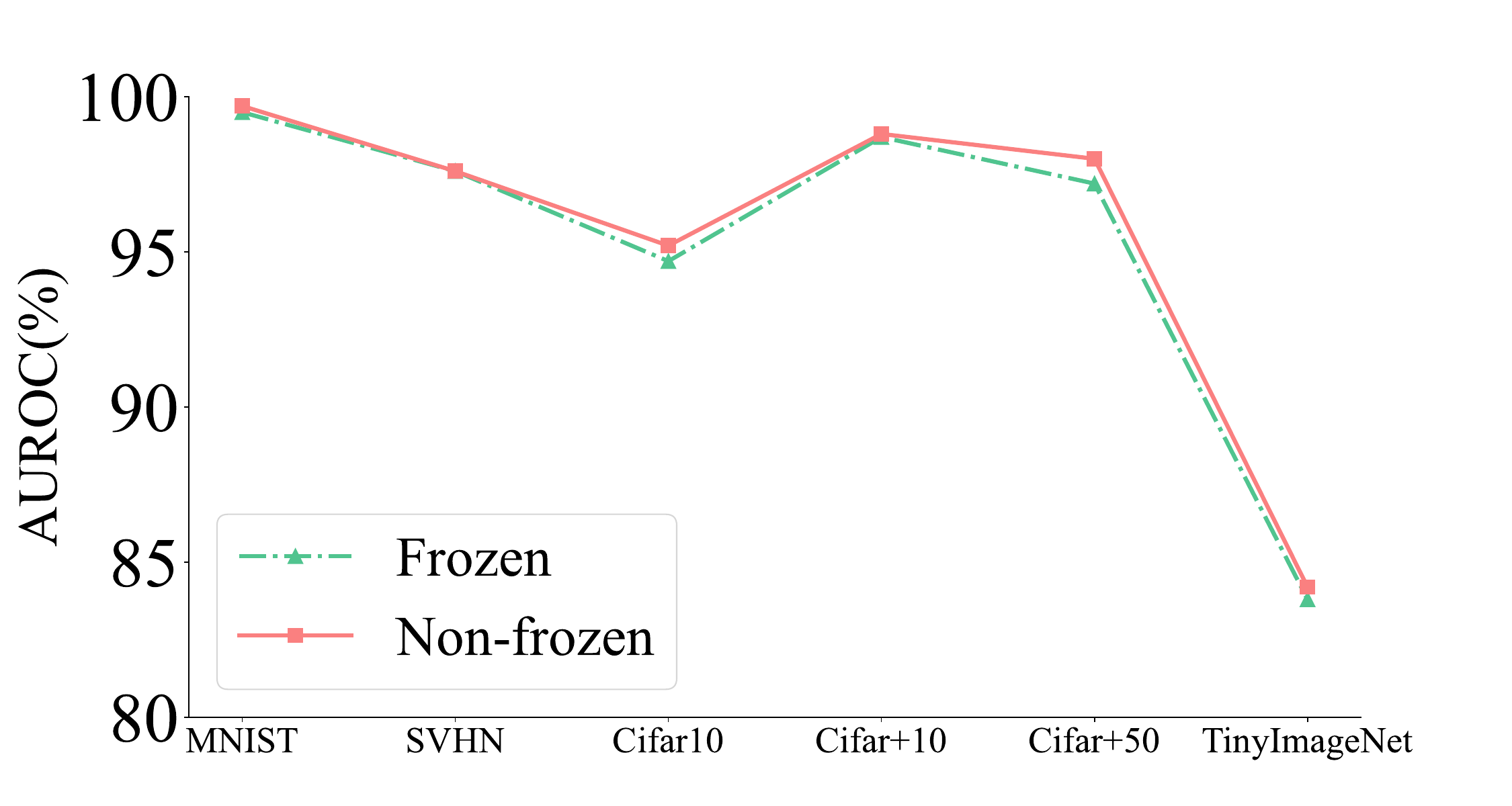}
\caption{Performance comparisons between frozen or non-frozen feature extractor in our OSSL on different benchmark datasets.}
\label{fig1:env}
\end{figure}

\subsection{Experiment on Frozen Feature Extractor}
Despite significant performance breakthroughs achieved by our OSSL, it is worth noting that its training process still requires fine-tuning of the feature extraction network $F(\cdot)$, which can significantly decrease the model's efficiency. One direct way to alleviate this issue is to freeze $F(\cdot)$. To evaluate the performance of OSSL in this situation, we conduct the related experiments and Figure 6 reports the results. We can find that the OSSL with frozen $F(\cdot)$ achieves comparable performance compared to the unfrozen counterpart on all benchmarks. This indicates that in the pursuit of efficiency but with less stringent performance requirements, the model can be trained by directly freezing the feature extractor well-trained on known class data.

\section{Conclusion}
In this paper, we rethink the OSR problem from the dynamic against dynamic perspective, where an open-set self-learning framework is proposed to construct the dynamic and changeable decision boundaries to adapt changing data distributions rather than a static and fixed ones in existing methods. Extensive experiments verify the effectiveness of our framework, refreshing the new performance records on almost all standard and cross-dataset benchmarks.

\appendix



\section*{Acknowledgments}
This research was supported in part by the National Natural Science Foundation of China (62106102, 62076124, 62376126), in
part by the Natural Science Foundation of Jiangsu Province (BK20210292), in part by the Hong Kong Scholars Program under Grant XJ2023035, in part by the RGC-GRF project RGC/HKBU12200820.

\bibliographystyle{named}
\bibliography{ijcai24}

\begin{thebibliography}{}

\bibitem[\protect\citeauthoryear{Bartler \bgroup \em et al.\egroup
  }{2022}]{bartler2022mt3}
Alexander Bartler, Andre B{\"u}hler, Felix Wiewel, Mario D{\"o}bler, and Bin
  Yang.
\newblock Mt3: Meta test-time training for self-supervised test-time adaption.
\newblock In {\em International Conference on Artificial Intelligence and
  Statistics}, pages 3080--3090. PMLR, 2022.

\bibitem[\protect\citeauthoryear{Bendale and
  Boult}{2015}]{Bendale2015TowardsOS}
Abhijit Bendale and Terrance~E. Boult.
\newblock Towards open set deep networks.
\newblock {\em 2016 IEEE Conference on Computer Vision and Pattern Recognition
  (CVPR)}, pages 1563--1572, 2015.

\bibitem[\protect\citeauthoryear{Cao \bgroup \em et al.\egroup
  }{2022}]{cao2022open}
Kaidi Cao, Maria Brbic, and Jure Leskovec.
\newblock Open-world semi-supervised learning.
\newblock {\em International Conference on Learning Representations}, 2022.

\bibitem[\protect\citeauthoryear{Cevikalp \bgroup \em et al.\egroup
  }{2023}]{cevikalp2023anomaly}
Hakan Cevikalp, Bedirhan Uzun, Yusuf Salk, Hasan Saribas, and Okan
  K{\"o}p{\"u}kl{\"u}.
\newblock From anomaly detection to open set recognition: Bridging the gap.
\newblock {\em Pattern Recognition}, 138:109385, 2023.

\bibitem[\protect\citeauthoryear{Chen \bgroup \em et al.\egroup
  }{2020}]{Chen2020LearningOS}
Guangyao Chen, Limeng Qiao, Yemin Shi, Peixi Peng, Jia Li, Tiejun Huang,
  Shiliang Pu, and Yonghong Tian.
\newblock Learning open set network with discriminative reciprocal points.
\newblock {\em ArXiv}, abs/2011.00178, 2020.

\bibitem[\protect\citeauthoryear{Chen \bgroup \em et al.\egroup
  }{2021}]{Chen2021AdversarialRP}
Guangyao Chen, Peixi Peng, Xiangqian Wang, and Yonghong Tian.
\newblock Adversarial reciprocal points learning for open set recognition.
\newblock {\em IEEE Transactions on Pattern Analysis and Machine Intelligence},
  44:8065--8081, 2021.

\bibitem[\protect\citeauthoryear{Deng \bgroup \em et al.\egroup
  }{2022}]{deng2022learning}
Shule Deng, Jin-Gang Yu, Zihao Wu, Hongxia Gao, Yansheng Li, and Yang Yang.
\newblock Learning relative feature displacement for few-shot open-set
  recognition.
\newblock {\em IEEE Transactions on Multimedia}, 2022.

\bibitem[\protect\citeauthoryear{Fang \bgroup \em et al.\egroup
  }{2021}]{fang2021learning}
Zhen Fang, Jie Lu, Anjin Liu, Feng Liu, and Guangquan Zhang.
\newblock Learning bounds for open-set learning.
\newblock In {\em International conference on machine learning}, pages
  3122--3132. PMLR, 2021.

\bibitem[\protect\citeauthoryear{Garg \bgroup \em et al.\egroup
  }{2023}]{Garg2023ComplementaryBO}
Saurabh Garg, Amrith~Rajagopal Setlur, Zachary~Chase Lipton, Sivaraman
  Balakrishnan, Virginia Smith, and Aditi Raghunathan.
\newblock Complementary benefits of contrastive learning and self-training
  under distribution shift.
\newblock {\em ArXiv}, abs/2312.03318, 2023.

\bibitem[\protect\citeauthoryear{Geng and Chen}{2022}]{geng2022collective}
Chuanxing Geng and Songcan Chen.
\newblock Collective decision for open set recognition.
\newblock {\em IEEE Transactions on Knowledge and Data Engineering},
  34(1):192--204, 2022.

\bibitem[\protect\citeauthoryear{Geng \bgroup \em et al.\egroup
  }{2020}]{geng2020guided}
Chuanxing Geng, Lue Tao, and Songcan Chen.
\newblock Guided cnn for generalized zero-shot and open-set recognition using
  visual and semantic prototypes.
\newblock {\em Pattern Recognition}, 102:107263, 2020.

\bibitem[\protect\citeauthoryear{Geng \bgroup \em et al.\egroup
  }{2021}]{geng2021recent}
Chuanxing Geng, Sheng-jun Huang, and Songcan Chen.
\newblock Recent advances in open set recognition: A survey.
\newblock {\em IEEE transactions on pattern analysis and machine intelligence},
  43(10):3614--3631, 2021.

\bibitem[\protect\citeauthoryear{Guo \bgroup \em et al.\egroup
  }{2021}]{Guo2021ConditionalVC}
Yunrui Guo, Guglielmo Camporese, Wenjing Yang, Alessandro Sperduti, and
  Lamberto Ballan.
\newblock Conditional variational capsule network for open set recognition.
\newblock {\em 2021 IEEE/CVF International Conference on Computer Vision
  (ICCV)}, pages 103--111, 2021.

\bibitem[\protect\citeauthoryear{Huang \bgroup \em et al.\egroup
  }{2022}]{huang2022class}
Hongzhi Huang, Yu~Wang, Qinghua Hu, and Ming-Ming Cheng.
\newblock Class-specific semantic reconstruction for open set recognition.
\newblock {\em IEEE transactions on pattern analysis and machine intelligence},
  45(4):4214--4228, 2022.

\bibitem[\protect\citeauthoryear{Iwasawa and Matsuo}{2021}]{iwasawa2021test}
Yusuke Iwasawa and Yutaka Matsuo.
\newblock Test-time classifier adjustment module for model-agnostic domain
  generalization.
\newblock {\em Advances in Neural Information Processing Systems},
  34:2427--2440, 2021.

\bibitem[\protect\citeauthoryear{Karisani}{2023}]{Karisani2023NeuralNA}
Payam Karisani.
\newblock Neural networks against (and for) self-training: Classification with
  small labeled and large unlabeled sets.
\newblock In {\em Annual Meeting of the Association for Computational
  Linguistics}, 2023.

\bibitem[\protect\citeauthoryear{Kong and Ramanan}{2021}]{kong2021opengan}
Shu Kong and Deva Ramanan.
\newblock Opengan: Open-set recognition via open data generation.
\newblock In {\em Proceedings of the IEEE/CVF International Conference on
  Computer Vision}, pages 813--822, 2021.

\bibitem[\protect\citeauthoryear{Krizhevsky}{2009}]{Krizhevsky2009}
Alex Krizhevsky.
\newblock Learning multiple layers of features from tiny images.
\newblock 2009.

\bibitem[\protect\citeauthoryear{Lake \bgroup \em et al.\egroup
  }{2015}]{Lake2015}
Brenden~M. Lake, Ruslan Salakhutdinov, and Joshua~B. Tenenbaum.
\newblock Human-level concept learning through probabilistic program induction.
\newblock {\em Science}, 350:1332 -- 1338, 2015.

\bibitem[\protect\citeauthoryear{Lee}{2013}]{Lee2013PseudoLabelT}
Dong-Hyun Lee.
\newblock Pseudo-label : The simple and efficient semi-supervised learning
  method for deep neural networks.
\newblock 2013.

\bibitem[\protect\citeauthoryear{Liu \bgroup \em et al.\egroup
  }{2021}]{liu2021ttt++}
Yuejiang Liu, Parth Kothari, Bastien Van~Delft, Baptiste Bellot-Gurlet, Taylor
  Mordan, and Alexandre Alahi.
\newblock Ttt++: When does self-supervised test-time training fail or thrive?
\newblock {\em Advances in Neural Information Processing Systems},
  34:21808--21820, 2021.

\bibitem[\protect\citeauthoryear{Lu \bgroup \em et al.\egroup
  }{2022}]{Lu2022PMALOS}
Jing Lu, Yunxu Xu, Hao Li, Zhanzhan Cheng, and Yi~Niu.
\newblock Pmal: Open set recognition via robust prototype mining.
\newblock In {\em AAAI Conference on Artificial Intelligence}, 2022.

\bibitem[\protect\citeauthoryear{Moon \bgroup \em et al.\egroup
  }{2022}]{Moon2022DifficultyAwareSF}
WonJun Moon, Junho Park, Hyun~Seok Seong, Cheol-Ho Cho, and Jae-Pil Heo.
\newblock Difficulty-aware simulator for open set recognition.
\newblock In {\em European Conference on Computer Vision}, 2022.

\bibitem[\protect\citeauthoryear{Mukherjee and
  Awadallah}{2020}]{mukherjee2020uncertainty}
Subhabrata Mukherjee and Ahmed~Hassan Awadallah.
\newblock Uncertainty-aware self-training for text classification with few
  labels.
\newblock {\em arXiv preprint arXiv:2006.15315}, 2020.

\bibitem[\protect\citeauthoryear{Neal \bgroup \em et al.\egroup
  }{2018}]{Neal2018OpenSL}
Lawrence Neal, Matthew~Lyle Olson, Xiaoli~Z. Fern, Weng-Keen Wong, and Fuxin
  Li.
\newblock Open set learning with counterfactual images.
\newblock In {\em European Conference on Computer Vision}, 2018.

\bibitem[\protect\citeauthoryear{Netzer \bgroup \em et al.\egroup
  }{2011}]{Netzer2011}
Yuval Netzer, Tao Wang, Adam Coates, A.~Bissacco, Bo~Wu, and A.~Ng.
\newblock Reading digits in natural images with unsupervised feature learning.
\newblock 2011.

\bibitem[\protect\citeauthoryear{Niu \bgroup \em et al.\egroup
  }{2022}]{niu2022efficient}
Shuaicheng Niu, Jiaxiang Wu, Yifan Zhang, Yaofo Chen, Shijian Zheng, Peilin
  Zhao, and Mingkui Tan.
\newblock Efficient test-time model adaptation without forgetting.
\newblock In {\em International conference on machine learning}, pages
  16888--16905. PMLR, 2022.

\bibitem[\protect\citeauthoryear{Oza and Patel}{2019}]{Oza2019C2AECC}
Poojan Oza and Vishal~M. Patel.
\newblock C2ae: Class conditioned auto-encoder for open-set recognition.
\newblock {\em 2019 IEEE/CVF Conference on Computer Vision and Pattern
  Recognition (CVPR)}, pages 2302--2311, 2019.

\bibitem[\protect\citeauthoryear{Perera and
  Patel}{2021}]{Perera2021GeometricTN}
Pramuditha Perera and Vishal~M. Patel.
\newblock Geometric transformation-based network ensemble for open-set
  recognition.
\newblock {\em 2021 IEEE International Conference on Multimedia and Expo
  (ICME)}, pages 1--6, 2021.

\bibitem[\protect\citeauthoryear{Russakovsky \bgroup \em et al.\egroup
  }{2014}]{Russakovsky2014ImageNetLS}
Olga Russakovsky, Jia Deng, Hao Su, Jonathan Krause, Sanjeev Satheesh, Sean Ma,
  Zhiheng Huang, Andrej Karpathy, Aditya Khosla, Michael~S. Bernstein,
  Alexander~C. Berg, and Li~Fei-Fei.
\newblock Imagenet large scale visual recognition challenge.
\newblock {\em International Journal of Computer Vision}, 115:211 -- 252, 2014.

\bibitem[\protect\citeauthoryear{Scheirer \bgroup \em et al.\egroup
  }{2013}]{Scheirer2013TowardOS}
Walter~J. Scheirer, Anderson Rocha, Archana Sapkota, and Terrance~E. Boult.
\newblock Toward open set recognition.
\newblock {\em IEEE Transactions on Pattern Analysis and Machine Intelligence},
  35:1757--1772, 2013.

\bibitem[\protect\citeauthoryear{Schneider \bgroup \em et al.\egroup
  }{2020}]{schneider2020improving}
Steffen Schneider, Evgenia Rusak, Luisa Eck, Oliver Bringmann, Wieland Brendel,
  and Matthias Bethge.
\newblock Improving robustness against common corruptions by covariate shift
  adaptation.
\newblock {\em Advances in neural information processing systems},
  33:11539--11551, 2020.

\bibitem[\protect\citeauthoryear{Scudder}{1965}]{ScudderProbabilityOE}
H.~J. Scudder.
\newblock Probability of error of some adaptive pattern-recognition machines.
\newblock {\em IEEE Trans. Inf. Theory}, 11:363--371, 1965.

\bibitem[\protect\citeauthoryear{Sun \bgroup \em et al.\egroup
  }{2020}]{sun2020test}
Yu~Sun, Xiaolong Wang, Zhuang Liu, John Miller, Alexei Efros, and Moritz Hardt.
\newblock Test-time training with self-supervision for generalization under
  distribution shifts.
\newblock In {\em International conference on machine learning}, pages
  9229--9248. PMLR, 2020.

\bibitem[\protect\citeauthoryear{Vaze \bgroup \em et al.\egroup
  }{2022}]{Vaze2022OpenSetRA}
Sagar Vaze, Kai Han, Andrea Vedaldi, and Andrew Zisserman.
\newblock Open-set recognition: A good closed-set classifier is all you need.
\newblock {\em the International Conference on Learning Representations},
  abs/2110.06207, 2022.

\bibitem[\protect\citeauthoryear{Wang \bgroup \em et al.\egroup
  }{2020}]{wang2020tent}
Dequan Wang, Evan Shelhamer, Shaoteng Liu, Bruno Olshausen, and Trevor Darrell.
\newblock Tent: Fully test-time adaptation by entropy minimization.
\newblock {\em arXiv preprint arXiv:2006.10726}, 2020.

\bibitem[\protect\citeauthoryear{Wang \bgroup \em et al.\egroup
  }{2022}]{wang2022openauc}
Zitai Wang, Qianqian Xu, Zhiyong Yang, Yuan He, Xiaochun Cao, and Qingming
  Huang.
\newblock Openauc: Towards auc-oriented open-set recognition.
\newblock {\em Advances in Neural Information Processing Systems},
  35:25033--25045, 2022.

\bibitem[\protect\citeauthoryear{Wei \bgroup \em et al.\egroup
  }{2021}]{wei2021crest}
Chen Wei, Kihyuk Sohn, Clayton Mellina, Alan Yuille, and Fan Yang.
\newblock Crest: A class-rebalancing self-training framework for imbalanced
  semi-supervised learning.
\newblock In {\em Proceedings of the IEEE/CVF conference on computer vision and
  pattern recognition}, pages 10857--10866, 2021.

\bibitem[\protect\citeauthoryear{Xie \bgroup \em et al.\egroup
  }{2020}]{xie2020self}
Qizhe Xie, Minh-Thang Luong, Eduard Hovy, and Quoc~V Le.
\newblock Self-training with noisy student improves imagenet classification.
\newblock In {\em Proceedings of the IEEE/CVF conference on computer vision and
  pattern recognition}, pages 10687--10698, 2020.

\bibitem[\protect\citeauthoryear{Xu \bgroup \em et al.\egroup
  }{2023}]{Xu2023ContrastiveOS}
Baile Xu, Furao Shen, and Jian Zhao.
\newblock Contrastive open set recognition.
\newblock In {\em AAAI Conference on Artificial Intelligence}, 2023.

\bibitem[\protect\citeauthoryear{Yoshihashi \bgroup \em et al.\egroup
  }{2018}]{Yoshihashi2018ClassificationReconstructionLF}
Ryota Yoshihashi, Wen Shao, Rei Kawakami, Shaodi You, Makoto Iida, and Takeshi
  Naemura.
\newblock Classification-reconstruction learning for open-set recognition.
\newblock {\em 2019 IEEE/CVF Conference on Computer Vision and Pattern
  Recognition (CVPR)}, pages 4011--4020, 2018.

\bibitem[\protect\citeauthoryear{You \bgroup \em et al.\egroup
  }{2019}]{You2019UniversalDA}
Kaichao You, Mingsheng Long, Zhangjie Cao, Jianmin Wang, and Michael~I. Jordan.
\newblock Universal domain adaptation.
\newblock {\em 2019 IEEE/CVF Conference on Computer Vision and Pattern
  Recognition (CVPR)}, pages 2715--2724, 2019.

\bibitem[\protect\citeauthoryear{Yu \bgroup \em et al.\egroup
  }{2015}]{Yu2015LSUNCO}
Fisher Yu, Yinda Zhang, Shuran Song, Ari Seff, and Jianxiong Xiao.
\newblock Lsun: Construction of a large-scale image dataset using deep learning
  with humans in the loop.
\newblock {\em ArXiv}, abs/1506.03365, 2015.

\bibitem[\protect\citeauthoryear{Zhang \bgroup \em et al.\egroup
  }{2020}]{Zhang2020HybridMF}
Hongjie Zhang, Ang Li, Jie Guo, and Yanwen Guo.
\newblock Hybrid models for open set recognition.
\newblock In {\em European Conference on Computer Vision}, 2020.

\bibitem[\protect\citeauthoryear{Zhang \bgroup \em et al.\egroup
  }{2022}]{zhang2022memo}
Marvin Zhang, Sergey Levine, and Chelsea Finn.
\newblock Memo: Test time robustness via adaptation and augmentation.
\newblock {\em Advances in Neural Information Processing Systems},
  35:38629--38642, 2022.

\bibitem[\protect\citeauthoryear{Zhou \bgroup \em et al.\egroup
  }{2021}]{zhou2021learning}
Da-Wei Zhou, Han-Jia Ye, and De-Chuan Zhan.
\newblock Learning placeholders for open-set recognition.
\newblock In {\em Proceedings of the IEEE/CVF conference on computer vision and
  pattern recognition}, pages 4401--4410, 2021.

\bibitem[\protect\citeauthoryear{Zhou}{2022}]{zhou2022open}
Zhi-Hua Zhou.
\newblock Open-environment machine learning.
\newblock {\em National Science Review}, 9(8):nwac123, 2022.

\bibitem[\protect\citeauthoryear{Zhuang \bgroup \em et al.\egroup
  }{2024}]{Zhuang2024CredibleTF}
Jingyu Zhuang, Kuo Wang, Liang Lin, and Guanbin Li.
\newblock Credible teacher for semi-supervised object detection in open scene.
\newblock 2024.

\bibitem[\protect\citeauthoryear{Zou \bgroup \em et al.\egroup
  }{2019}]{zou2019confidence}
Yang Zou, Zhiding Yu, Xiaofeng Liu, BVK Kumar, and Jinsong Wang.
\newblock Confidence regularized self-training.
\newblock In {\em Proceedings of the IEEE/CVF international conference on
  computer vision}, pages 5982--5991, 2019.

\end{thebibliography}

\end{document}